\documentclass[letterpaper]{article} 
\usepackage[draft]{aaai25}  
\usepackage{times}  
\usepackage{helvet}  
\usepackage{courier}  
\usepackage[hyphens]{url}  
\usepackage{graphicx} 
\usepackage{natbib}  
\usepackage{caption} 
\usepackage{algorithm}
\usepackage{algorithmic}
\usepackage{graphicx}
\usepackage{subfigure}
\usepackage{amsfonts}
\usepackage{multirow}
\usepackage{amssymb}
\usepackage{newtxmath}
\usepackage{url}
\usepackage{bm}
\usepackage{marvosym}
\usepackage{enumitem}
\usepackage{booktabs}       
\usepackage{nicefrac}       
\usepackage{microtype}      
\usepackage{color}
\usepackage{multicol}
\usepackage{algorithm}
\usepackage{algorithmic}
\usepackage{amssymb}
\usepackage[table]{xcolor}

\usepackage{newfloat}
\usepackage{listings}
\DeclareCaptionStyle{ruled}{labelfont=normalfont,labelsep=colon,strut=off} 
\floatstyle{ruled}
\newfloat{listing}{tb}{lst}{}
\floatname{listing}{Listing}
\title{Conditional Prototype Rectification Prompt Learning}
\author{
    Haoxing Chen\textsuperscript{\rm 1}, Yaohui Li\textsuperscript{\rm 2}, Zizheng Huang\textsuperscript{\rm 1,2}, Yan Hong\textsuperscript{\rm 1}, Zhuoer Xu\textsuperscript{\rm 1}\\ Zhangxuan Gu\textsuperscript{\rm 1},
    Jun Lan\textsuperscript{\rm 1}, Huijia Zhu\textsuperscript{\rm 1}, Weiqiang Wang\textsuperscript{\rm 1}
}
\affiliations{
    \textsuperscript{\rm 1} Tiansuan Lab, AntGroup\\
    \textsuperscript{\rm 2} Nanjing University\\
    hx.chen@hotmail.com
}

\usepackage{bibentry}

\begin{document}

\maketitle

\begin{abstract}
Pre-trained large-scale vision-language models (VLMs) have acquired profound understanding of general visual concepts. Recent advancements in efficient transfer learning (ETL) have shown remarkable success in fine-tuning VLMs within the scenario of limited data, introducing only a few parameters to harness task-specific insights from VLMs. Despite significant progress, current leading ETL methods tend to overfit the narrow distributions of base classes seen during training and encounter two primary challenges: (i) only utilizing uni-modal information to modeling task-specific knowledge; and (ii) using costly and time-consuming methods to supplement knowledge. To address these issues, we propose a Conditional Prototype Rectification Prompt Learning (CPR) method to correct the bias of base examples and augment limited data in an effective way. Specifically, we alleviate overfitting on base classes from two aspects. First, each input image acquires knowledge from both textual and visual prototypes, and then generates sample-conditional text tokens. Second, we extract utilizable knowledge from unlabeled data to further refine the prototypes. These two strategies mitigate biases stemming from base classes, yielding a more effective classifier. Extensive experiments on 11 benchmark datasets show that our CPR achieves state-of-the-art performance on both few-shot classification and base-to-new generalization tasks.  Our code is avaliable at \url{https://github.com/chenhaoxing/CPR}.
\end{abstract}
\section{Introduction}
\begin{figure}[t]
\centering
\includegraphics[width=0.8\linewidth]{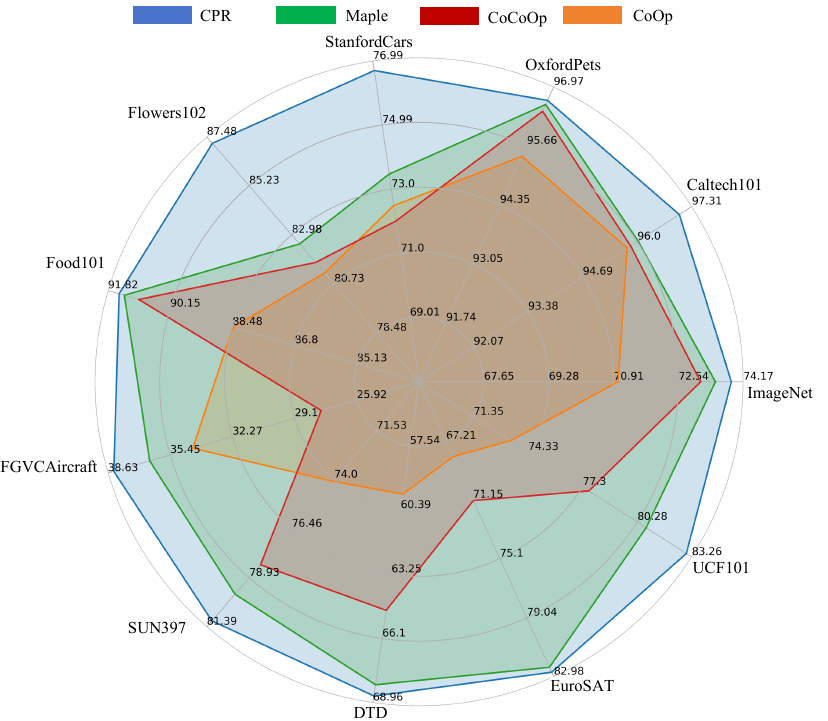}
  \caption{Comparison of different efficient transfer learning methods on base-to-new generalization task.}
  \label{fig1}
\end{figure}

    Leveraging the vast array of image-text association pairs, the trained visual-language models (VLMs) embody critical general knowledge, thereby exhibiting enhanced generalizability across diverse tasks. In recent developments, several visual-language models like Contrastive Language-Image Pre-training~\cite{CLIP} (CLIP), Flamingo~\cite{Flamingo}, and ALIGN~\cite{ALIGN} have been introduced. While VLMs are proficient at extracting visual and textual descriptions, their training necessitates extensive, high-quality datasets. Nevertheless, amassing substantial data for training task-specific models in real-world visual-language applications poses a challenge. To circumvent this issue, efficient transfer learning (ETL) methods~\cite{CoOp,KgCoOp,CoCoOp,GraphAdapter,APE}have been devised to fine-tune pre-trained VLMs for downstream tasks, demonstrating remarkable efficacy in various few-shot or zero-shot visual recognition tasks.

There are two notable ETL methods for VLMs: prompt tuning~\cite{CoOp,KgCoOp,CoCoOp,prograd} and adapter-style tuning~\cite{Clip-adapter,GraphAdapter,APE}. Prompt tuning modifies the textual classifier by incorporating learnable prompts at the input, significantly outperforming zero-shot CLIP with few-shot examples, seen in CoOp~\cite{CoOp} and CoCoOp~\cite{CoCoOp}. Conversely, adapter-style tuning adjusts textual classifiers or visual features at the output, using the textual encoder just once. For example, CLIP-Adapter~\cite{Clip-adapter} employs a bottleneck layer to fine-tune embeddings, surpassing zero-shot CLIP by 3.02\% on ImageNet in a one-shot scenario. TaskRes~\cite{teskres} adjusts textual embeddings with task-specific parameters, acting as independent residuals. Another approach enhances downstream task knowledge using large scale databases or generative models, like Sus-X~\cite{SUS-X} and C2A~\cite{roy2023Cap2Aug}.

However, two main limitations exist in current ETL research: 1) Many approaches, like CLIP-Adapter~\cite{Clip-adapter} and TaskRes~\cite{teskres}, focus on task-specific knowledge from only one modality, relying solely on visual or textual features for adaptation. 2) Using generative models or large scale datasets to augment datasets typically requires substantial computational resources, impacting efficiency.
In low-data scenarios, limited samples struggle to adequately reveal structural knowledge for downstream tasks, potentially biasing the model toward certain attributes like color or shape. This can limit transferability and generalization. Therefore, integrating multi-modality knowledge and discovering efficient strategies to enrich the dataset are essential for VLM tuning, addressing the biases that stem from data scarcity.

To alleviate these limitations, we propose two prototype rectification strategies, referred to as conditional prototype rectification learning, which aim to model task-specific knowledge for downstream tasks by integrating textual and visual structural knowledge as well as knowledge from unlabeled samples.
To tackle the first limitation, we introduce the Conditional Adapter (CoAdapter), which leverages the input image alongside visual and textual prototypes to capture structured knowledge pertinent to downstream tasks, producing sample-specific prototypes. This equips the feature adapter with a comprehensive and accurate understanding essential for these tasks.
Addressing the second limitation, we devise the Nearest Neighbor Rectification (NNR) strategy. It identifies the $k$ nearest samples from the unlabeled dataset that align with the prototypes generated by the CoAdapter, facilitating rectification. This process effectively utilizes the latent knowledge within the unlabeled samples, enriching the model's learning context.
As shown in Figure 1, CPR achieves significant improvements over the previous state-of-the-art algorithms on generalization tasks.

The contributions of this paper are summarized as follows:
\begin{itemize}
    \item We propose conditional prototype rectification learning (CPR), which introduce two prototype rectification strategies designed to amend the bias inherited from base examples and effectively augment limited data.
    \item We introduce a novel adapter-style tuning approach, the Conditional Adapter (CoAdapter), for efficiently transfer learning of vision-language models. This strategy harnesses dual-modality structure knowledge from both textual and visual domains, allowing the feature adapter to utilize integrated visual and textual insights. This fusion facilitates enhanced learning of task-specific knowledge from downstream tasks, leading to effective VLM tuning.
    \item we propose the Nearest Neighbor Rectification (NNR), a method of knowledge completion that addresses the issue of biased knowledge without the need for auxiliary or synthetic data. NNR capitalizes on the concept of nearest neighbors to extract valuable insights from test samples, enriching the information available from a few shots.
    \item We evaluate our CPR across 11 popular benchmarks in both few-shot classification and base-to-new generalization settings. The evaluations reveal that CPR markedly surpasses earlier prompt-based and adapter-style methodologies, showcasing notable efficacy even in demanding fine-grained image classification tasks, exemplified by Caltetch101.
\end{itemize}

\begin{figure*}[t]
\centering
\includegraphics[width=0.9\linewidth]{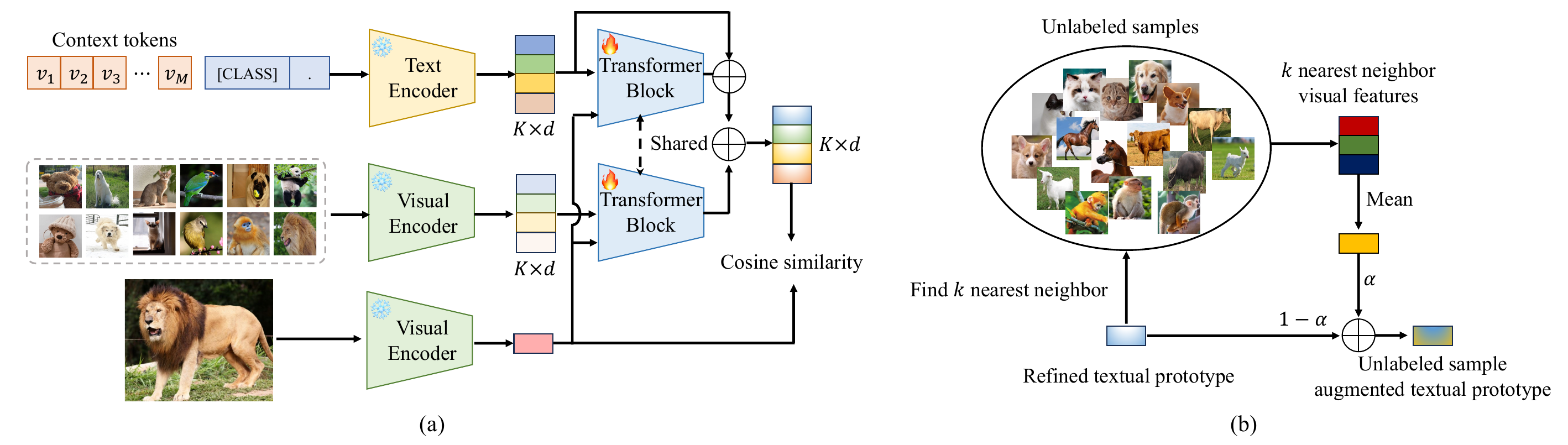}
\caption{Our approach, Conditional Prototype Rectification Prompt Learning (CPR), includes two strategies: (a) Conditional Adapter exploits connections between input images and both visual and textual prototypes, integrating textual and visual structures to model task-specific knowledge. (b) Nearest Neighbor Rectification leverages unlabeled data to avoid external or synthetic data, addressing biases and limitations in few-shot learning.}
\label{CPR}
\end{figure*}

\section{Related Works}
\subsection{Visual-Language Pre-training}
The principal aim of Visual-Language Pre-training is to execute multimodal pre-training employing extensive datasets comprising image-text pairs, aiming to develop universal representations for text and images. This approach is intended to improve the effectiveness of model in various downstream tasks. These tasks notably include zero-shot learning, few-shot classification, cross-modal generation, open-world segmentation, and open-world detection, among others. In this context, models such as VisualBERT~\cite{VisualBERT}, OSCAR~\cite{Oscar}, and Uniter~\cite{UNITER} use BERT~\cite{BERT} to process raw text, achieving significant results in multimodal tasks like Visual Question Answering (VQA), which demonstrates their substantial application potential. These successes emphasize the importance of well-designed fusion encoders in merging cross-modal interactions. Additionally, recent studies like CLIP~\cite{CLIP}, DeCLIP~\cite{DeCLIP}, and ALIGN~\cite{ALIGN} have demonstrated the effectiveness of visual-language contrastive learning in producing features that are useful for downstream tasks. These features are acquired by simply calculating the dot product between visual and linguistic embeddings, offering a straightforward understanding of multimodal interactions. Moreover, these approaches avoid the need for self-attention or cross-attention mechanisms, allowing for the precomputation and storage of multimodal embeddings without the need for further modifications, thus increasing efficiency and easing their integration into other tasks.

\subsection{Efficient Transfer Learning}
Efficient Transfer Learning aims to transfer task knowledge to downstream tasks by adjusting parameters in pre-trained visual-language models. ETL methods include prompt tuning and adapter-style tuning.

Prompt tuning uses task-related textual tokens to infer task-specific knowledge, exemplified by the text template of CLIP for zero-shot prediction. Context Optimization (CoOp)~\cite{CoOp} uses learnable soft prompts from few-shot examples, but generates fixed prompts for each task, ignoring image feature variation. Conditional Context Optimization (CoCoOp)~\cite{CoCoOp} creates image-specific prompts using a lightweight neural network. ProDA~\cite{ProDA}  learns a prior distribution for prompts, and ProGrad~\cite{prograd} selectively updates prompt gradients aligned with "general knowledge."

Adapter-style tuning modulates textual and visual features for task adaptation. CLIP-Adapter~\cite{Clip-adapter} uses a bottleneck layer for few-shot classification, and TaskRes~\cite{teskres} introduces a task-independent adapter.

In existing methods, the works most relevant to our study are CoCoOp and Sus-X~\cite{SUS-X}. CoCoOp can be regarded as the baseline model for our proposed CPR. Unlike CoCoOp, our CPR considers not only the input image itself but also the structural relationship between the input image and both visual and textual prototypes to further enhance the performance of learnable prompts. Sus-X introduces a text-image generative model to augment the original support set, whereas CPR utilizes unlabeled samples from the test set for prototype refinement, offering greater efficiency.

\section{Preliminaries}
We first provide brief reviews on the adopted VLM, i.e., contrastive language-image pre-training (CLIP)~\cite{CLIP}, recap two mainstream approaches for ETL on VLMs, i.e., prompt tuning and adapter-style tuning.
\subsection{Contrastive Language-Image Pre-training}
Contrastive Language-Image Pre-training~\cite{CLIP} aims to acquire visual representations through natural language supervision. CLIP consists of two encoders: one for images and another for text. The image encoder can be either ResNet~\cite{ResNet} or Vision Transformer~\cite{ViT} (ViT), transforming images into feature vectors. The text encoder is a Transformer~\cite{transformer} that takes a sequence of word tokens as input and produces a vectorized representation.

CLIP is trained using 400 million image-text pairs and employs a contrastive loss to learn a joint embedding space for both modalities, enabling CLIP to effectively capture a wide range of visual concepts and learn universal visual representations.
Specifically, for an input image $x$ belonging to one of the classes $Y = \{y_1, y_2, ..., y_C\}$, the image encoder $f$ extracts image features $z = f(x)\in\mathbb{R}^d$.
To obtain the corresponding text features $w_i \in \{1,...,C\}$, names of all given classes are filled into the fixed prompt template "a photo of a [CLASS]" (e.g., bear, urocissa erythrorhyncha), generating text descriptions $A$, which are then further encoded into text embeddings $W = g(A) \in \mathbb{R}^{d\times C}$. Image-text alignment is optimized based on the cosine similarity of features.

During the testing phase, CLIP is capable of classifying a query image $x$ into $C$ possible categories. This is accomplished by computing the cosine similarity between the query image embedding vector and the set of text embeddings $W$. The prediction probability is then determined using the formula:
\begin{equation}
    p(y|z) = \frac{{\rm exp}({\rm sim}({z, W_y})/\tau)}{\sum_{i=1}^{C}{\rm exp}({\rm sim}({z, W_i})/\tau)},
\end{equation}
where ${\rm sim}(\cdot,\cdot)$denotes the cosine similarity, and $\tau$ is a learned temperature parameter. 

\subsection{Revisiting Previous Efficient Transfer Paradigms}
Motivated by the effectiveness of Efficient Transfer Learning (ETL) techniques in natural language processing, such as prompt tuning and Adapter, recent developments like CoCoOp and CLIP-Adapter have adapted these concepts for use in visual-language models (VLMs).	

CoOp~\cite{CoOp} introduces prompt tuning to VLMs for the first time, utilizing $M$ learnable context vectors $v = \{v_1,v_2,...,v_M\}$to generate task-specific templates, replacing the fixed text prompt templates used in CLIP.
Specifically, $t_i = \{v_1,v_2,...,v_M, c_i\}$ integrates learnable context vectors $v$ with the class token embedding $c_i$. This combined prompt is then input into the text encoder $g(\cdot)$. During training process, CoOp maintains the pre-trained VLMs parameters unchanged and solely adjusts the learnable vectors $v$.  
CoCoOp~\cite{CoCoOp} is an enhanced version of CoOp.
CoOp suggests that contexts conditioned on individual instances offer superior generalization because they pivot attention from a narrow set of classes (thus reducing overfitting risks) toward each unique input instance, facilitating broader task applicability. Meanwhile, CoCoOp introduces a Meta-Net to produce a conditional token that is subsequently amalgamated with the context vectors, enhancing adaptability and specificity to each instance.

Adapter-style tuning integrates extra modules 
$\phi_{\omega}(\cdot)$ into pre-trained models to modify the pre-trained features $f$ into new features $f'$. Generally, adapter-style tuning can be expressed as:
\begin{equation}
    f' = f + \alpha \phi_{\omega}(f)
\end{equation}
where $\alpha$ is a scaling factor.In CLIP-Adapter, the adapter module $\phi_{\omega}$ consists of two linear transformation layers with a ReLU activation function situated between them. CLIP-Adapter explores the use of adapters for both visual and textual components, applying the adapter module separately to the image and text branches of CLIP. This investigation reveals that both types of adapters—visual and textual—yield similarly effective performance. When training on downstream tasks, adapter-style approaches like CLIP-Adapter focus on tuning the parameters of these adapter modules only, leaving the foundational pre-trained CLIP model parameters unchanged. This strategy allows for task-specific customization while maintaining the general capabilities of the underlying model. Moreover, recent developments in SuS-X~\cite{SUS-X} and C2A~\cite{roy2023Cap2Aug} have utilized generative diffusion models or large-scale external databases to achieve enhanced feature representations $f'$.

\section{Approach}
In this section, we will provide a detailed description of the two prototype rectification strategies, i.e., conditinal adapter and nearest neighbor rectification. The whole framework of our coditional prototype rectification prompt learning is depicted in Figure 2, which is composed of two strategy, including conditional adapter and nearest neighbor rectification.

\subsection{Conditional Adapter}
Previous works like CoCoOp~\cite{CoCoOp} considered how input images impact text prototypes. And CLIP-Adapter~\cite{Clip-adapter}, TaskRes~\cite{teskres}, and Tip-Adapter~\cite{Tip-Adapter} derive features using separate visual or textual features. In contrast, we propose conditional adapter (CoAdapter), which leverages the connections between input images and both visual and textual prototypes, modeling sample-specific knowledge for downstream tasks through the integrated textual and visual structural knowledge.

As shown in Figure \ref{CPR} (a), the conditional adapter is developed based on CoOp~\cite{CoOp}. For a given input image $x$, we utilize the feature $z$ of the input image and the relationship between the image prototype $\mathbf{V}$/text prototype $\mathbf{W}$ to generate a sample-specific residual.
This entails processing both the feature $z$ and the prototypes through a transformer block, which then produces the sample-specific residual.
Specifically, we apply the image feature as query and the prototype embeddings as both key and value into an attention block.
Then, the visual residual $R_v \in \mathbb{R}^{d\times C}$ is calculated as an exemplar of the sample-specific residual using the following set of equations:
\begin{gather}
    Q_v, K_v, V_v = F_q(z), F_k(\mathbf{V}), F_v(\mathbf{V}),\\
    R_v = {\rm Norm}({\rm FFN_0}({\rm Softmax}(\frac{Q_v\cdot K_v^\top}{\sqrt{d}})V_v)+{\rm FFN_1}(z)),
\end{gather}
where ${\rm Norm}$ denotes the layer normalization process. The $F_q$, $F_k$ and $F_v$ represent the multilayer perceptron layers that generate the query, key, and value for the attention mechanism, respectively. The terms ${\rm FFN_0}$ and ${\rm FFN_1}$ are also multilayer perceptrons involved in further processing the attention output and the original input feature $z$ before combining them. By using a similar computation method to visual residuals, we can obtain the textual residual $R_t$. Afterward, the fused features $P = \mathbf{W} + R_t + R_v$ are aggregated and used as refined textual prototype for classification, effectively combining insights from both the visual and textual domains to improve the performance on specific tasks.

\begin{figure*}[t]
	\centering
	\includegraphics[width=0.99\linewidth]{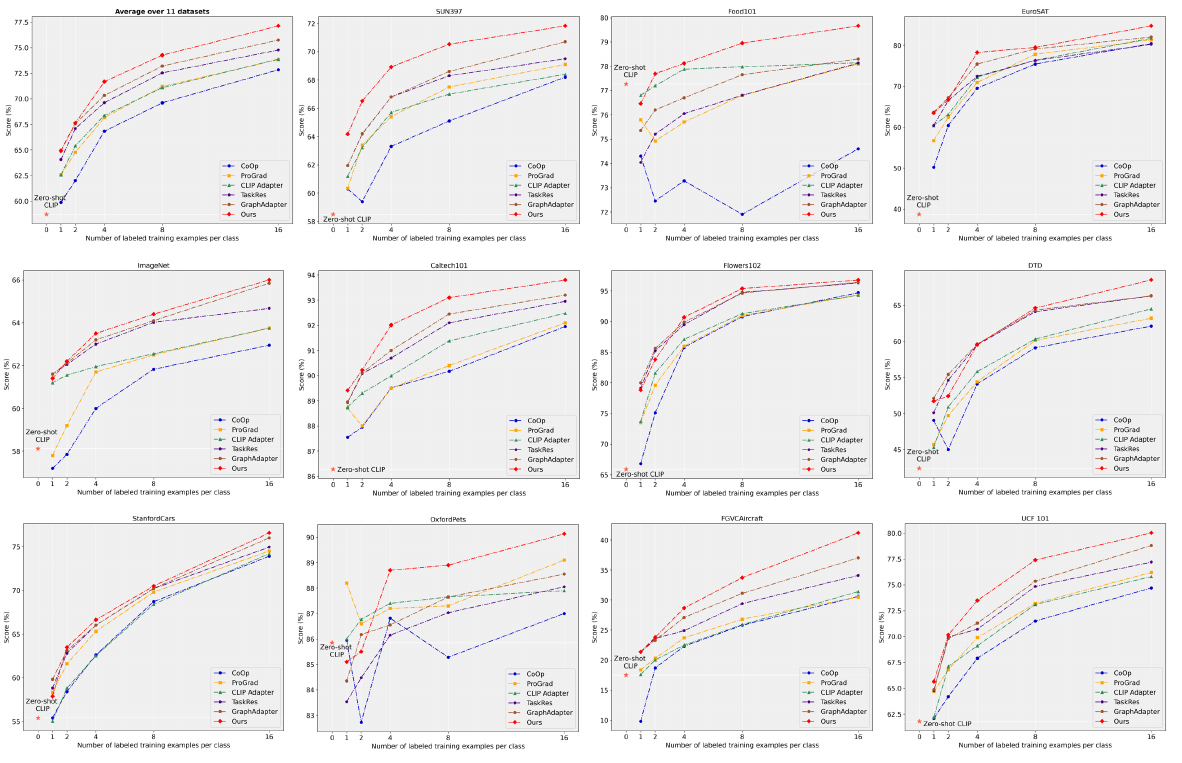}
	\caption{The performance comparison for few-shot learning (1-/2-/4-/8-/16-shot) is conducted on 11 benchmark datasets, with the top-left indicating the averaged accuracy across these datasets.}
	\label{CPR-few}
\end{figure*}

\subsection{Nearest Neighbor Rectification}
Inspired by C2A \cite{roy2023Cap2Aug} and SuS-X \cite{SUS-X} which enhance prototypes using generative diffusion models~\cite{sd,Instructpix2pix,diffute} or expansive external databases, our work seeks to enrich the limited knowledge obtained from few-shot samples. We aim to leverage the potential of unlabeled samples, circumventing the need for auxiliary or synthetic data, to mitigate the inherent biases and constraints associated with few-shot learning scenarios. 

Due to data scarcity, the features $P$ we derive might display considerable bias. To address this, we employ features extracted from high-confidence unlabeled images, which are utilized to refine and adjust the feature representation $P$, enhancing its robustness and generalizability. Specifically, for each class prototype $P_i$, we find its $k$ nearest neighbors within the unlabeled dataset and utilize their features to recalibrate $P_i$. This adjustment process can be mathematically articulated as follows:
\begin{equation}
    P'_i = \alpha\cdot P_i + (1-\alpha)\cdot \frac{\sum_{j=1}^k z^u_j}{k},
\end{equation}
where $z^u_j$ is the $j$-th nearest neighbor of the $K$ nearest neighbors, and $\alpha$ is the hyper-parameter to adjust the weights of $P$ and knowledge from unlabeled samples.

\subsection{Tuning for Downstream Tasks}
To ensure that the fused features $P$ we obtain do not significantly deviate from those generated by the pre-trained CLIP model, and to enhance the model's generalization ability, we employ a consistency constraint. This constraint encourages the fused features to remain similar to the representations from the original pre-trained model, which has learned robust and generalizable features from a vast amount of data. We use a pre-trained large language model (LLM) to generate more descriptive sentences from template textual inputs, which are then used to guide the fused features. By doing so, we ensure that the fused features have a high degree of similarity with the textually-augmented representations provided by the LLM. We can denote the consistency constraint as:
\begin{equation}
    \mathcal{L}_{cons} = \frac{1}{C}\sum_{i=1}^C(1 - \frac{P_i\cdot g(t_i)}{||P_i||\cdot||g(t_i)||})
\end{equation}

During training, we only update the 
learnable context vectors $v$ together with the transformer block in CoAdapter, maintaining the base classifier and the image branch of CLIP unchanged. Given an image,  supervised classification loss is computed as:
\begin{equation}
    \mathcal{L}_{cls} = -log\frac{{\rm exp}({\rm sim}({z, P'_y})/\tau)}{\sum_{i=1}^{C}{\rm exp}({\rm sim}({z, P'_i})/\tau)}.
\end{equation}
Adding both losses with a balancing factor $\lambda$, we get the final loss function of CPR is:
\begin{equation}
\mathcal{L} = \mathcal{L}_{cls} + \lambda \cdot \mathcal{L}_{cons}.
\end{equation}

\begin{table}[htbp]
  \centering
  \resizebox{0.48\textwidth}{!}{
\begin{tabular}{cc|ccccccc|c}
\hline
\multirow{2}{*}{Dataset} &      & CLIP & CoOp & CoCoOp  &KgCoOp& MaPLe &PromptSRC &TCP& CPR \\
&&\textcolor[rgb]{0.5, 0.5, 0.5}{(ICML'21)}&\textcolor[rgb]{0.5, 0.5, 0.5}{ (IJCV'22)}&\textcolor[rgb]{0.5, 0.5, 0.5}{ (CVPR'22)}&\textcolor[rgb]{0.5, 0.5, 0.5}{ (CVPR'23)}&\textcolor[rgb]{0.5, 0.5, 0.5}{ (CVPR'23)}&\textcolor[rgb]{0.5, 0.5, 0.5}{ (ICCV'23)}&\textcolor[rgb]{0.5, 0.5, 0.5}{ (CVPR'24)}&
\\
\midrule
&Base&69.34  & 82.63&80.47 &   80.73&82.28& 84.12 &\textbf{84.13}&84.02  \\
Average &New&74.22  &67.99& 71.69&  73.60& 75.14& 75.02 &75.36&\textbf{76.22}   \\
&HM &71.70  &74.60  &75.83&  77.00& 78.55& 79.31  &79.51& \textbf{79.93}
\\
\midrule
&Base&72.43&76.46&75.98& 75.83&76.66& \textbf{77.75} &77.27& 77.31 \\
ImageNet &New&68.14& 66.31 & 70.43 & 69.96& 70.54&70.70 &69.87& \textbf{70.72} \\
&HM & 70.22&71.02& 73.10   & 72.78& 73.47&\textbf{74.06}  &73.38& 73.87 \\
\midrule
&Base& 96.84&98.11& 97.96  &97.72&97.74& 98.13  &98.23&\textbf{98.64}  \\
Caltech101 &New&94.00&93.52& 93.81&  94.39& 94.36&93.90  &94.67& \textbf{95.43}  \\
&HM &95.40&95.76& 95.84&  96.03  & 96.02 & 95.97  &96.62& \textbf{97.01}  \\
\midrule
&Base&91.17&94.24&95.20 &94.65 &95.43 &95.50  &94.67&\textbf{95.58}  \\
OxfordPets &New& 97.26& 96.66& 97.69 & 97.76 & 97.76 &97.40  &97.20&\textbf{97.79} \\
&HM &94.12& 95.43& 96.43 & 96.18 & 96.58 & 96.44  &95.92& \textbf{96.67} \\
\midrule
&Base&63.37 &76.20&70.49 &71.76&72.94 &78.40&\textbf{80.80}&78.32  \\
StanfordCars &New& 74.89& 69.14& 73.59 & 75.04& 74.00& 74.73 &74.13& \textbf{75.12}  \\
&HM &68.65& 72.49& 72.01 & 73.36& 73.47 &75.52 &\textbf{77.32}& 76.69 \\
\midrule
&Base&72.08&97.63&94.87 &95.00&95.92 &\textbf{97.90} &97.73&97.15  \\
Flowers102 &New&77.80&69.55& 71.75 & 74.73& 72.46&76.77  &75.57& \textbf{79.07}  \\
&HM & 74.83&81.23& 81.71 & 83.65& 82.56  &86.06 &85.23 & \textbf{87.18}   \\
\midrule
&Base&90.10&89.44&90.70&90.50&90.71&90.63&90.57&\textbf{90.95}\\
Food101 &New& 91.22& 87.50& 91.29 & 91.70& 92.05 & 91.50&91.37& \textbf{92.09}   \\
&HM & 90.66&88.46& 90.99  & 91.09& 91.38 & 91.06 &90.97& \textbf{91.52}   \\
\midrule
&Base&27.19 &39.24& 33.41 &36.21&37.44 &\textbf{42.30} &41.97&41.26 \\
FGVCAircraft &New& 36.29& 30.49& 23.71 & 33.55& 35.61&\textbf{36.97}&34.43& 35.78 \\
&HM & 31.09& 34.30& 27.74  & 34.83 &36.50&\textbf{39.46}&37.83& 38.33 \\
\midrule
&Base&69.36&80.85& 79.74 &80.29&80.82&\textbf{82.83} &82.63& 82.77 \\
SUN397 &New& 75.35& 68.34& 76.86  & 76.53 & 78.70 &79.00  &78.20& \textbf{79.48}\\
&HM &72.23& 74.07&  78.27 & 78.36 & 79.75 &80.87 &80.35& \textbf{81.09}  \\
\midrule
&Base&53.24 &80.17& 77.01 &77.55 &80.36 & 82.60  &82.77& \textbf{82.95} \\
DTD &New& 59.90 &47.54& 56.00  & 54.99 & \textbf{59.18}& 57.50 &58.07&58.57 \\
&HM &56.37&59.68& 64.85  & 64.35& 68.16& 67.80&68.25& \textbf{68.66}\\
\midrule
&Base&56.48&91.54& 87.49  &85.64&94.07& 92.40 &91.63&\textbf{92.93} \\
EuroSAT &New& 64.05& 54.44&  60.04  & 64.34& 73.23&68.43&\textbf{74.73}&74.57   \\
&HM & 60.03& 68.27& 71.21&   73.48 & 82.35&78.63 &82.32& \textbf{82.68} \\
\midrule
&Base&70.53&85.14& 82.33 & 82.89&83.00& 86.93&\textbf{87.13}&86.35 
 \\
UCF101 &New& 77.50& 64.47& 73.45 & 76.67& 78.66& 78.33&\textbf{80.77}&79.83  \\
&HM & 73.85& 73.37& 77.64 &  79.65& 80.77&82.41 &\textbf{83.83}& 82.96 \\
\bottomrule
\end{tabular}}
\caption{Comparison with existing methods in the base-to-new generalization setting with ViT-B/16 as the backbone. The context length M is 4 for prompot-based methods with the 16-shots samples from the base classes.}
\end{table}

\section{Experiment}

\subsection{Datasets and Implementation Details}
Similar to CoCoOp~\cite{CoCoOp} and PromptSRC~\cite{PromptSRC}, our evaluation of the proposed method encompasses two primary settings: 1) generalization from base to new classes within a dataset; and 2) few-shot image classification. These evaluations are conducted leveraging the pretrained CLIP~\cite{CLIP} model, ensuring a consistent basis for comparison and thorough assessment of generalization and adaptability capabilities. More detailed results will be given in the Supplementary materials.

\textbf{Datasets}. Following CoOp~\cite{CoOp}, CoCoOp~\cite{CoCoOp}, and TCP~\cite{TCP}, the base-to-new generaliation and few-shot image classification are conducted on 11 image classification datasets, i.e., ImageNet~\cite{Imagenet} and Caltech101~\cite{Caltech101} for generic object classification; OxfordPets~\cite{oxfordpets}, StanfordCars~\cite{stanfordcars}, Flowers102~\cite{flowers102}, Food101~\cite{FOOD101}, and FGVCAircraft~\cite{FGVC-aircraft} for fine-grained visual categorization, EuroSAT~\cite{Eurosat} for satellite image  classification, UCF101~\cite{UCF101} for action recognization, DTD~\cite{DTD} for texture classification, and SUN397~\cite{SUN397} for scene recognition. 

\textbf{Training details}. Our experiments are executed using the same frameworks as established in TCP~\cite{TCP} and GraphAdapter~\cite{GraphAdapter}, incorporating the CLIP model as the core component. The experiments utilize ResNet-50~\cite{ResNet} and ViT-B/16~\cite{ViT} as vision backbones, aligning with the setups in CoOp where the context length is consistently held at 16. To ensure statistical reliability and fairness in comparison, the reported performance metrics are the averages derived from three distinct random seeds. 
Consistency with the original studies also extends to the training scheme; we adhere to the same training epochs, training schedule, and data augmentation strategies as described in CoOp~\cite{CoOp}.
Regarding hyperparameters, we set the $\alpha$ for the base-to-new generalization and the few-shot learning setups to 8.0 and 1.0, respectively, and uniformly set $\alpha$ to 0.95, and we adjust the number of neighbors $k$ to 5.
The computational groundwork for all experiments is provided by a single NVIDIA RTX A100 80G GPU.

\begin{table}[t]
    \centering
    \resizebox{\linewidth}{!}{
    \begin{tabular}{ccccc|c}
    \toprule
    Metric & Baseline & w/o CoAdapter& w/o NNR & w/o $\mathcal{L}_{cons}$ & CPR \\
    \midrule
    Base &82.63&83.15&83.07&83.17&\textbf{84.02} \\
    New &67.99&72.05&71.86&73.35&\textbf{76.22} \\\midrule
    H &74.60&77.20&77.06&77.95&\textbf{79.93} \\
    \bottomrule
    \end{tabular}}
    \caption{Ablation studies.}
\end{table}
\begin{table}[t]
    \centering \resizebox{\linewidth}{!}{
    \begin{tabular}{cccccc}
    \toprule
    Model&1-shot&2-shot&4-shot&8-shot&16-shot\\
    \midrule
    CoAdapter-I&16.55&20.76&24.04&28.23&32.34\\
    CoAdapter-T&19.92&22.23&24.72&28.26&32.10\\
    \midrule
    CoAdapter&\textbf{21.36}&\textbf{23.79}&\textbf{28.66}&\textbf{33.72}&\textbf{41.17}\\
\bottomrule
    \end{tabular}}
    \caption{Performance comparison of three CoAdapter variants on FGVCAircraft under few-shot learning setting.}
\end{table}

\subsection{Few-shot Classification}
The results of our experiments are illustrated in Figure \ref{CPR-few}, demonstrating the superior performance of our CPR method relative to previous ETL approaches across varying shot evaluations (1-/2-/4-/8-/16-shot) when averaged over 11 benchmark datasets. In particular, during the 16-shot evaluation, CPR attains an average success rate of 77.12\%, surpassing GraphAdapter by 1.38\% and TaskRes by 2.37\%. It is worth noting that CPR maintains its lead in performance even in the demanding FGVCAircraft dataset, which focuses on fine-grained classification, outstripping GraphAdapter by 4.17\% in the 16-shot setting. Our CPR can achieve these promising results because CPR can alleviate overfitting by exploiting the structure knowledge with the dual prototypes and using knowledge from unlabeled data.
\subsection{Generalization From Base-to-New Classes}
Similar to previous works~\cite{TCP,PromptSRC,MaPLe}, we divide each dataset into two categories: base classes and new classes. Mirroring the zero-shot setting, the new classes are disjoint from the base classes. To assess the generalizability of the CoOp-based approaches, all evaluated methods, including the proposed CPR, utilize the base classes for tuning and perform evaluations on the new classes. The specific results are detailed in Table 1, which provides comprehensive performance metrics across all 11 datasets using the ViT-B/16 backbone with 16-shot samples.

As shwon in Table 1, CPR method surpasses existing methodologies in achieving a higher average performance based on the harmonic mean across all settings, underlining its enhanced capability for generalizing from base to new classes. Specifically, CPR demonstrates superior performance on new classes in seven out of the eleven datasets, including ImageNet, Caltech101, OxfordPets, StanfordCars, Flowers102, Food101 and SUN397. The exceptional performance highlights the improvement of CPR in generalization to new classes compared to the current methods. Moreover, CPR consistently achieves top results for the base classes in the majority of scenarios. Hence, CPR effectively enhances generalizability to new classes while maintaining or even improving base class performance, thereby securing the highest Harmonic mean across all evaluated datasets.

\subsection{Ablation Studies}
\textbf{The effects of each prototype rectification strategy in CPR.} To evaluate the effectiveness of the proposed strategy, ablation studies were conducted. As shown in Table 2, the results indicate that both strategies significantly enhanced the generalizability of model and mitigated overfitting relative to the base class. 
If the CoAdapter is removed from CPR, the accuracy on new class recognition tasks would decrease by 5.5\%. Similarly, if NNR is not used, the accuracy would decrease by 5.7\%.
These research findings confirm that each strategy can independently enhance recognition performance, and their synergistic application is particularly effective in improving accuracy in efficient transfer learning scenarios.

\begin{figure}[t]
\centering
\includegraphics[width=0.9\linewidth]{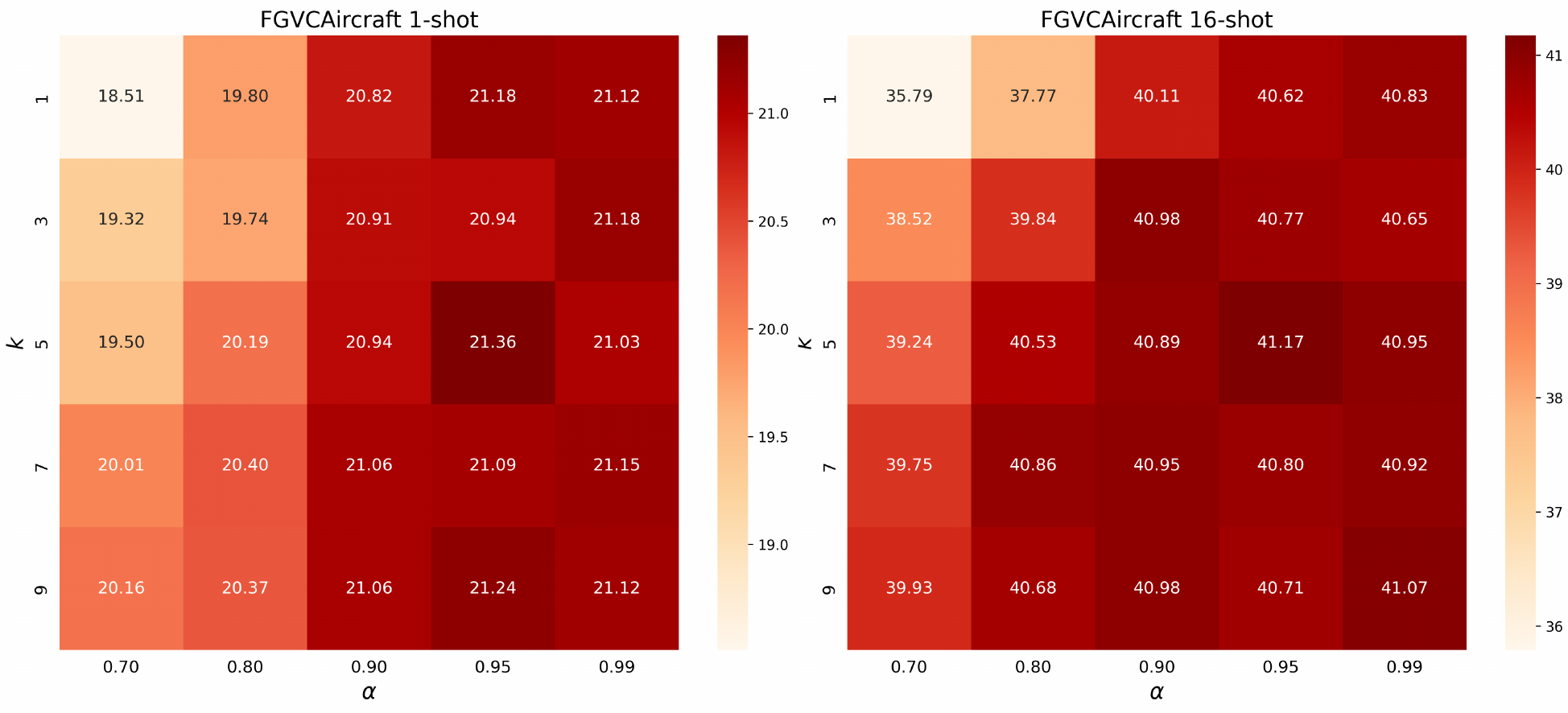}
\caption{The ablation studies for two coefficients $\alpha$ and $k$ on FGVCAircraft under few-shot learning setting.}
\label{CPR}
\end{figure}

\noindent
\textbf{The effects of dual prototype in CoAdapter.} The distinctiveness of our CoAdapter lies in its utilization of dual prototypes to generate sample-conditioned text tokens. Therefore, we investigate the effectiveness of another two variants of our GraphAdapter, i.e., CoAdapter-I, and CoAdapter-T. CoAdapter-I leverages only the visual prototype for generating sample-conditioned text tokens, whereas CoAdapter-T employs solely the textual prototype for this purpose. As shown in Table 3, the CoAdapter exhibits superior performance when it simultaneously leverages the prototypes from both modalities. This enhancement can be attributed to the complementary nature of the information provided by each modality: the visual modality offers details about object appearance and structure, while the textual modality provides semantic and contextual insights. The integration of these diverse information streams enables the CoAdapter to develop more comprehensive and robust data representations, thereby improving its generalization capabilities and performance on downstream tasks.

\begin{table}[t]
    \centering
    \resizebox{0.6\linewidth}{!}{
    \begin{tabular}{c|c|cc|c}
    \toprule
    Backbone&Model&Base&New&H\\
    \midrule
    \multirow{2}{*}{ResNet-50}&CoOp&27.01&24.48&25.68\\
    &CPR&\textbf{28.27}&\textbf{26.51}&\textbf{27.36}\\\midrule
    \multirow{2}{*}{ResNet-101}&CoOp&31.75&25.97&28.57\\
    &CPR&\textbf{33.15}&\textbf{29.33}&\textbf{31.12}\\\midrule
    \multirow{2}{*}{ViT-B/32}&CoOp&30.49&25.67&27.88\\
    &CPR&\textbf{30.92}&\textbf{29.55}&\textbf{30.22}\\\midrule
    \multirow{2}{*}{ViT-B/16}&CoOp&39.27& 30.52& 34.35 \\
    &CPR&\textbf{41.26}&\textbf{35.78}&\textbf{38.33}\\
\bottomrule
    \end{tabular}}
    \caption{Comparisons different backbones on FGVCAircraft under base-to-new generalization setting.}
\end{table}

\noindent
\textbf{The effects of neighbors in NNR.} 
In the NNR strategy, determining the optimal value of $k$—the number of nearest unlabeled samples to associate with each fused prototype $P_i$-is essential for knowledge completion across each class. To ascertain the most appropriate $k$ value, we conducted an ablation study within a few-shot learning framework, evaluating $k$ values from the set $\{1, 3, 5, 7, 9\}$. The results, depicted in Figure 4, reveal that the model attains its best performance when $k$ is set to 5. This outcome implies that, within the context of our experiments, integrating information from five nearest neighbors achieves the optimal balance between acquiring sufficient knowledge and mitigating the influence of irrelevant information, thereby enhancing the model's predictive accuracy most effectively.

\noindent
\textbf{The effects of $\alpha$ in NNR.} 
In the NNR strategy, the parameter $\alpha$ plays a pivotal role in determining the extent to which the model benefits from the knowledge in unlabeled data. To identify the ideal setting for $\alpha$, we carried out a series of ablation experiments. As shown in Figure 4, the results indicate that the model achieves optimal performance when $\alpha$ is set to 0.95. This outcome suggests that an appropriate $\alpha$ value can effectively balance the insights from unlabeled data with the requirements to ensure the reliability of the correction process. In contrast, too high of an $\alpha$ value might not fully utilize the embedded knowledge in the unlabeled data, while too low of an $\alpha$ might introduce unnecessary noise or irrelevant details, potentially impairing the performance.

\noindent
\textbf{The effects of different backbones.} As shown in Table 4, we also evaluate the effectiveness of our CPR method across various CLIP visual backbones, namely ResNet-50, ResNet-101, ViT-B/32, and ViT-B/16. Across all four visual backbones, CPR consistently outperforms prior methods, demonstrating its robustness and adaptability to different underlying architectures. This result highlights CPR's versatility and its ability to leverage distinct visual features extracted by various backbones, thus affirming its broad applicability and effectiveness in enhancing model performance.

\noindent
\textbf{The effects of $\lambda$.} To investigate the effectiveness of hyperparameter $\lambda$, as specified in Equation (8) in Section Approch, we conduct experiments on FGVCAircraft by varying different $\lambda$.
$\mathcal{L}_{cons}$ aims to enhance the generalizability for unseen classes by minimizing the distance between learnable fused features and fixed textual enhanced representations provided by the LLM
As shown in Table 5, selecting an optimal value for $\lambda$ is crucial for enhancing model performance. Specifically, for base-to-new generalization, setting $\lambda$ to 8 yields the most beneficial outcome. An appropriate $\lambda$ can avoid learning text tokens with poor generalization on base classes while generating sample-specific representations. On the other hand, for few-shot learning setting, the optimal performance is observed when $\lambda$ is adjusted to 1.

\noindent
\textbf{Limitations.} The limitation of our CPR is rooted in the selection process of unlabeled data. While the $k$-nearest neighbor algorithm provides a foundational approach to augmenting our dataset, it lacks a definitive level of confidence, potentially limiting the quality of the knowledge integration. We posit that a more precise and sophisticated selection method could significantly enhance the quality of the supplementary knowledge, thereby improving the performance on downstream tasks.

\begin{table}[t]
    \centering\resizebox{\linewidth}{!}{
    \begin{tabular}{cccc|ccccc}
    \toprule
    \multirow{2}{*}{$\lambda$}  &\multicolumn{3}{c}{Base-to-New} &\multicolumn{5}{c}{Few-shot Learning}\\\cline{2-4} \cline{5-9}
    &Base&New&H&1-shot&2-shot&4-shot&8-shot&16-shot\\
    \midrule
    0.0&38.78&34.83&36.69&19.11&22.38&26.79&32.85&41.03\\
    0.1&39.02&34.01&36.34&19.25&23.32&27.65&33.65&40.97\\
    1.0&42.80&33.95&37.87&\textbf{21.36}& \textbf{23.79}& \textbf{28.66}& \textbf{33.72}&\textbf{41.17}\\
    2.0&41.18&32.03&36.03&21.05&23.12&28.42&33.22&40.86\\
    5.0&40.88&35.33&37.90&20.88&22.99&27.97&33.12&40.54\\
    8.0&\textbf{41.26}& \textbf{35.78}& \textbf{38.33}&20.99&22.76& 27.88&33.21&40.76\\
    10.0&39.44&35.31&37.26&20.64&22.54&27.75&333.04 &40.74\\
\bottomrule
    \end{tabular}}
    \caption{The ablation studies for coefficient $\lambda$ on FGVCAircraft.}
\end{table}

\section{Conclusion}
In this paper, we provide an in-depth examination of the constraints faced by previous approaches in scenarios characterized by scarce data, identifying two main limitations: 1) the reliance on a single modality for modeling task-specific knowledge, and 2) the employment of resource-intensive and protracted techniques for knowledge enhancement, which are crucial for tasks requiring data efficiency. To address these issues, we introduce an innovative Conditional Prototype Rectification Prompt Learning (CPR) methodology for visual-language models. CPR aims to rectify biases present in base examples and effectively enrich limited datasets. It achieves this by amalgamating fused textual and visual prototype knowledge with insights gleaned from unlabeled samples, thereby enabling the classifier to adapt more effectively to downstream tasks. Our comprehensive testing across 11 benchmark datasets demonstrates the effectiveness of our CPR on few-shot learning and generalization tasks.

\bibliography{aaai25}

\end{document}